\documentclass[10pt,twocolumn,letterpaper]{article}

\usepackage{cvpr}              
\definecolor{cvprblue}{rgb}{0.21,0.49,0.74}
\usepackage[pagebackref,breaklinks,colorlinks,allcolors=cvprblue]{hyperref}
\usepackage{multirow,colortbl}
\usepackage[ruled,vlined]{algorithm2e}

\def\paperID{416} 
\def\confName{CVPR}
\def\confYear{2026}

\title{SAGE: Style-Adaptive Generalization for Privacy-Constrained\\Semantic Segmentation Across Domains}
\author{Qingmei Li$^1$, Yang Zhang$^2$, Peifeng Zhang$^2$, Haohuan Fu$^{1,3}$\thanks{Corresponding authors.}, Juepeng Zheng$^{2,3 *}$\\
$^1$Tsinghua Shenzhen International Graduate School\quad 
$^2$Sun Yat-Sen University\quad \\
$^3$National Supercomputing Center in Shenzhen\\
{\tt\small qingmeili@sz.tsinghua.edu.cn, haohuan@tsinghua.edu.cn, zhengjp8@mail.sysu.edu.cn}
}

\begin{document}
\maketitle
\begin{abstract}
Domain generalization for semantic segmentation aims to mitigate the degradation in model performance caused by domain shifts. However, in many real-world scenarios, we are unable to access the model parameters and architectural details due to privacy concerns and security constraints. Traditional fine-tuning or adaptation is hindered, leading to the demand for input-level strategies that can enhance generalization without modifying model weights. To this end, we propose a \textbf{S}tyle-\textbf{A}daptive \textbf{GE}neralization framework (\textbf{SAGE}), which improves the generalization of frozen models under privacy constraints. SAGE learns to synthesize visual prompts that implicitly align feature distributions across styles instead of directly fine-tuning the backbone. Specifically, we first utilize style transfer to construct a diverse style representation of the source domain, thereby learning a set of style characteristics that can cover a wide range of visual features. Then, the model adaptively fuses these style cues according to the visual context of each input, forming a dynamic prompt that harmonizes the image appearance without touching the interior of the model. Through this closed-loop design, SAGE effectively bridges the gap between frozen model invariance and the diversity of unseen domains. Extensive experiments on five benchmark datasets demonstrate that SAGE achieves competitive or superior performance compared to state-of-the-art methods under privacy constraints and outperforms full fine-tuning baselines in all settings.
\end{abstract}    
\section{Introduction}
\label{sec:intro}

Semantic segmentation is a fundamental task in computer vision \cite{ji2024segment,pan2022label,lu2023scaling}, with applications spanning from autonomous driving \cite{Li2023MSeg3D} to medical image analysis \cite{xiao2024convolutional,zhang2024optimization}. With the rapid advancement of deep learning techniques, semantic segmentation has witnessed remarkable progress in recent years \cite{zhou2024image,cuarunta2025heavy}. Notable approaches such as FCNs \cite{long2015fully,chen2018encoder}, RNN-based method \cite{byeon2015scene}, the DeepLab series \cite{chen2017deeplab}, and Transformer-based method \cite{cheng2021per,liang2023local} have achieved strong performance on standard benchmarks. However, these methods typically assume that the source and target domain are drawn from the same underlying distribution \cite{li2024hyunida}. In real-world scenarios, this assumption rarely holds due to variations in urban
landscape, weather, and lighting conditions\cite{chen2022learning,sakaridis2021acdc,cao2025exploring}. Domain shifts can lead to significant performance degradation when deployed on unseen target domains \cite{li2025boosting}.

\begin{figure*}[t]
  \centering
  \includegraphics[width=\textwidth]{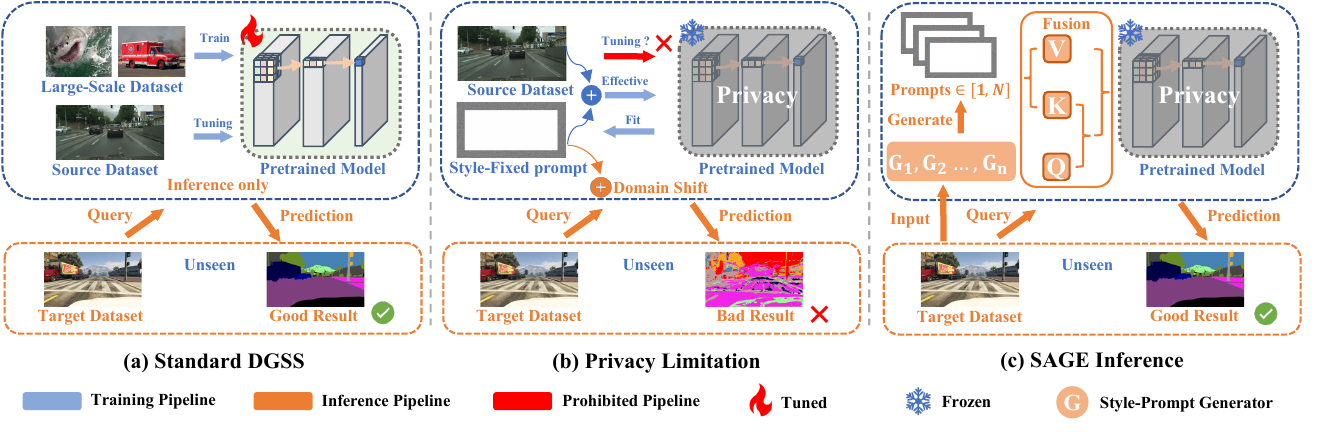}
  \vspace{-1em}
  \caption{Overview of domain generalization settings under privacy constraints. (a) Standard DGSS requires backbone tuning. (b) Style-fixed prompting is ineffective on unseen domains. (c) SAGE inference overcomes privacy limitations.}
  \label{fig:overview}
  \vspace{-0.5em}
\end{figure*}

To address this issue, domain generalized semantic segmentation (DGSS) has emerged, which aims to train a model solely on the source domain while ensuring strong generalization to unseen target domains \cite{bi2024learning,benigmim2024collaborating}. The central goal is to enhance the model's robustness to domain shifts through either style decoupling \cite{ISW,peng2022semantic, peng2021global} or style augmentation \cite{huang2023style,lee2022wildnet,zhao2022style}. As shown in Figure \ref{fig:overview} (a), owing to the powerful feature extraction capabilities of foundation models such as Transformers \cite{vaswani2017attention}, most DGSS approaches prefer to fine-tune backbones pre-trained on large-scale datasets (\textit{e.g.}, ImageNet), rather than training models from scratch, which is often computationally prohibitive \cite{peng2022semantic,ISW,yi2024learning}. 
However, existing DGSS methods overlook a crucial real-world constraint. In practical deployments, segmentation models are frozen due to privacy protection, intellectual property, or deployment security, making their internal parameters inaccessible for fine-tuning or adaptation. The situation raises a critical practical problem: \textit{How can segmentation models generalize across domains without accessing model internals?}.

The first challenge is the \textbf{inaccessibility of model parameters} due to privacy concerns or security constraints \cite{shokri2015privacy,chen2024federated,el2022differential}. The backbone is kept frozen in IP-protected, encrypted, or sandboxed deployments, while gradients w.r.t. the input are available during training. In the privacy-aware environments, parameter tuning becomes infeasible, rendering conventional DGSS methods ineffective. Inspired by the success of visual prompting in natural language and vision tasks \cite{jia2022visual}, recent efforts have explored steering model behavior by attaching learnable prompts to the input. However, most existing prompting strategies assume parameter accessibility and thus operate on internal model layers \cite{guo2024x,chen2024rsprompter}, which violates the privacy-aware constraint.

External visual prompts, which are applied directly to the input, offer a promising direction to guide model behavior without accessing internal parameters \cite{bahng2022exploring,huang2023diversity}. While some recent works adopt this strategy, they typically train on a single domain and fail to handle the large appearance variations that arise across unseen environments \cite{yu2024a2xp}. As a result, the learned prompt becomes style-fixed, as shown in Figure \ref{fig:overview} (b), it is only effective on data whose style resembles the source domain. The second challenge emerges: the \textbf{high diversity of target domain styles}, which renders a single style-fixed prompt ineffective. Due to the large visual diversity in target domains and the inability to access target data during training, a single fixed prompt cannot generalize well across domains. Furthermore, since the prompt is only used for inference on target domain, there is no opportunity to further optimize the prompt during testing.

To address these challenges, we explore \textbf{\textit{SAGE}} (\textbf{\textit{S}}tyle-\textbf{\textit{A}}daptive \textbf{\textit{GE}}neralization), a privacy-aware generalization framework designed to improve the generalization of frozen segmentation models without accessing internal parameters. The key of SAGE is to implicitly align feature distributions across diverse styles through input-level optimization, enabling robustness under unseen domains. First, we augment the source domain using the style transfer strategy and train multiple style-prompt generators, each conditioned on a distinct stylized variant of the source data, effectively working around the parameter inaccessibility constraints of the model. During inference, the learned cues are dynamically composed via an attention-based mechanism into an adaptive visual prompt. The instance-wise adaptation enables the model to generate the most suitable prompt, addressing the challenge of high style diversity in target domains. The main contributions are summarized as follows.

$\bullet$ We highlight for the first time in DGSS the necessity of privacy model generalization, addressing the increasingly common scenario where deployed models are frozen for security concerns.


$\bullet$ We propose SAGE, which achieves input-level adaptation by learning diverse style prompts from stylized source data and dynamically fusing them based on input context, thereby enabling robust generalization without accessing model parameters.


$\bullet$ We conduct comprehensive experiments on five benchmark datasets, indicating that SAGE achieves competitive or superior performance compared to state-of-the-art methods in three settings under privacy constraints and outperforms full fine-tuning baselines in all settings.

\section{Related works}
\label{rel}

\subsection{Domain Generalization} 
Domain generalization (DG) aims to improve a model’s ability to generalize to unseen target domains without accessing target domain data during training \cite{zheng2022multisource,benigmim2024collaborating}. Since style discrepancies are a major cause of domain shift, existing DG methods for semantic segmentation can be grouped into style decoupling and style augmentation approaches.
Style decoupling aims to suppress domain-specific styles in feature representations. For example, SW \cite{pan2019switchable} used a combination of Batch Normalization (BN) and Instance Normalization (IN) to balance discriminative power and style removal. ISW \cite{ISW} introduced a whitening loss to eliminate style-related covariance. Peng et al. \cite{peng2021global} randomized global and local textures to reduce overfitting to source patterns. DIRL \cite{xu2022dirl} used domain sensitivity as prior knowledge to guide feature recalibration and whitening.
In contrast, style augmentation can enhance source diversity by encouraging the model to learn style-agnostic features across a wide range of appearances. Peng et al. \cite{peng2022semantic} translated images into various painting styles. Yue et al. \cite{yue2019domain} randomized synthetic image styles using auxiliary datasets. SiamDoGe \cite{wu2022siamdoge} avoided auxiliary domains by using self-guided randomization and consistency training.
Our method aligns with the style augmentation paradigm. By applying style transfer using an auxiliary dataset, we train multiple prompt generators conditioned on diverse stylized views of the source domain, enabling better coverage of target domain variations.

\subsection{Visual Prompt Tuning} 
Visual prompt tuning (VPT) aims to adapt pre-trained models to downstream tasks by fitting a small number of additional parameters without modifying the original model weights. VPT \cite{jia2022visual} first introduced this concept and inspired further methods like X-Prompt \cite{guo2024x} and SHIP \cite{zhu2024semantic}, which enrich prompt expressiveness via semantic structures or multimodal information. However, both VPT and the aforementioned methods typically require access to and modification of the model architecture, which is infeasible in our black-box setting. In contrast, Bahng et~al.\cite{bahng2022exploring} explored applying visual prompts as direct perturbations to the input image, thereby adapting frozen pre-trained models without modifying token embeddings or adding trainable components within the model. Building on this, Huang et al.\cite{huang2023diversity} proposed DAM-VP, which introduces a diversity-aware strategy and meta-learning to enhance visual prompting, especially on datasets with high intra-domain diversity. Yu et al.\cite{yu2024a2xp} further improved adaptation performance by combining expert adaptation and attention-based generalization.
Yet, these methods often use style-fixed prompts, which limits adaptability under diverse target domains. Therefore, we design an instance-aware mechanism that dynamically selects and fuses prompts based on each input image's style characteristics.

\subsection{Attention Mechanism} 
The core idea of the attention mechanism is to enhance important features while suppressing less relevant ones. From the original Transformer architecture \cite{vaswani2017attention}, to lightweight modules like Squeeze-and-excitation networks (SE) \cite{hu2018squeeze}, and deeper designs such as Residual Attention Networks \cite{wang2017residual}, attention is widely adopted for improving representation learning.
Depending on the sources of Query, Key, and Value matrices, attention can be categorized into self-attention (the same source) and cross-attention (different sources). For example, A2XP \cite{yu2024a2xp} uses cross-attention to integrate multiple expert prompts from diverse domains. X-prompt \cite{guo2024x} computes both spatial and channel attention over RGB and X-modality patch tokens, which are then fused into spatial-modal attention weights to generate a complementary multi-modal prompt.
Inspired by these ideas, we adopt the cross-attention mechanism to organically fuse style-prompts generated by multiple style-prompt generators, and integrate the fused prompt with the input image. This allows it to dynamically produce an adaptive prompt that best aligns with the style of the query image.
\section{Method}
\label{method}

\subsection{Preliminary}
We consider the problem of improving the performance of a frozen pre-trained segmentation model $\Phi$ on unseen target domains $\mathcal{D}_t$ without modifying its internal parameters or architecture. Given a labeled source domain $\mathcal{D}_s = \{(x_i,y_i)\}$, where $x_i$ is the input image and $y_i$ is the corresponding segmentation map, the challenge is to enable $\Phi$ to generalize to $\mathcal{D}_t$ despite significant style and distribution gaps. This task is particularly challenging when differences in visual appearance lead to degraded segmentation results $\hat{y}$, such as noisy boundaries or missing objects.

To tackle this challenge, we propose SAGE, a privacy-aware generalization framework consisting of \textit{style-prompt generation} and \textit{adaptive prompt fusion}. First, we employ style transfer techniques to construct $n$ stylized variants of the source dataset and train a dedicated prompt generator for each. These generators learn to produce adaptive style prompts tailored to different visual appearances. When these prompts are attached to the input images, they can boost the segmentation performance of the frozen model. Furthermore, to avoid the impractical manual style assignment, we pass the original source images through all $n$ generators and utilize a lightweight cross-attention module to fuse the resulting prompts into a unified representation. The fused prompt adaptively reflects the input style and, once integrated with the image, enables better generalization to unseen domains.



\begin{figure*}[t]
  \centering
  \includegraphics[width=\textwidth]{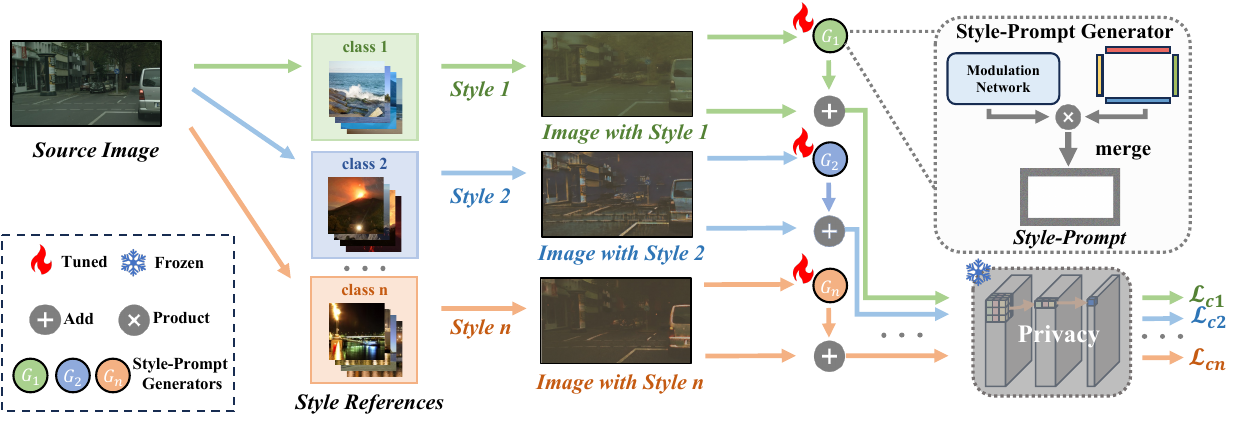}
  \caption{The training process of SPG: each style reference guides the source image through a dedicated generator $G_i$ to produce a style-aware prompt. Prompts are merged and used to optimize a frozen privacy model. Each style-prompt generator consists of a modulation network and a learnable prompt template, producing the final style-prompt via element-wise modulation.}
  \label{fig:SPG}
\end{figure*}

\subsection{Style-Prompt Generation}
Since the unseen target domain often contains a wide range of everyday scenes with visual styles that differ significantly from the source domain, it is critical to generate effective prompts tailored for these diverse style. To this end, we randomly sample $n$ categories from ImageNet \cite{deng2009imagenet}, which is called style references in our framework. For each reference style, we use image-to-image translation \cite{zhu2017unpaired} to stylize the source images accordingly. Figure \ref{fig:SPG} illustrates the overall architecture of the Style-Prompt Generation (SPG) phase. For more details on the style transfer process, please refer to the Appendix \ref{style_transfer}.

However, even within the same visual style, differences in objects and scenes lead to variations in the regions of interest for the prompt. Therefore, a fixed prompt per style is insufficient. To address this, we design a style-prompt generator that dynamically produces prompts based on the content of each image.
Concretely, each generator $G_i$ for style $i \in \{1,2,\ldots,n\}$ maintains a learnable prompt template $\mathcal{T}_i = \{P_t, P_b, P_l, P_r\}$, where the learnable parameters are spatially constrained to the border regions, while the central area is filled with zeros. This border-shaped structure is designed to efficiently encode general prior knowledge associated with each style.
Given an input image $\mathbf{X} \in \mathbb{R}^{B \times C \times H \times W}$, a lightweight modulation network $\mathcal{M}$ predicts attention coefficients to adaptively reweight each semantic border of the template.
The modulation network $\mathcal{M}$ consists of a series of residual blocks, followed by global feature aggregation. Each ModulatorBlock is defined as:
\begin{equation}
    f_\text{block}(\mathbf{X}; \theta_i) = \mathcal{F}(\mathbf{X}; \theta_i) + \mathcal{S}(\mathbf{X}; \phi_i)
\end{equation}
where $\mathcal{F}(\cdot)$ represents the main convolution path, $\mathcal{S}(\cdot)$ represents the shortcut connection, and $\theta_i, \phi_i$ are the learnable parameters. Specifically, for each block:
\begin{equation}
    \mathcal{F}(\mathbf{X}; \theta_i) = \text{BN}_2(\text{Conv}_2(\sigma(\text{BN}_1(\text{Conv}_1(\mathbf{X})))))
\end{equation}

The input image is first processed by a cascaded series of Modulator Blocks, where the channel dimension is progressively expanded and the spatial resolution is reduced accordingly, resulting in semantically rich feature representations. These features are intended to capture high-level, style-relevant discriminative information.
\begin{equation}
    \mathbf{X}_{mod} = f_{mod}(\mathbf{X};\theta) \in \mathbb{R}^{B\times D\times \frac{H}{8} \times \frac{W}{8}}
\end{equation}
where $f_{mod}$ denotes the composition of several ModulatorBlocks, and $\theta$ is the set of all block parameters. Detailed configurations are provided in Appendix \ref{theoSPG}.
Next, the extracted features are projected to a certain dimension via a 1×1 convolution, followed by global pooling to obtain a compact representation. This representation is then reorganized into four sets of boundary modulation coefficients $\alpha = [\alpha_p, \alpha_b, \alpha_l, \alpha_r]$. 
Each of the four coefficient sets is then multiplied with its corresponding semantic region, enabling content-aware modulation of the border prompts. 
\begin{equation}
\begin{aligned}
    P'_t = P_t \odot \alpha_{t}, \quad
    P'_b = P_b \odot \alpha_{b}, \\
    P'_l = P_l \odot \alpha_{l}, \quad
    P'_r = P_r \odot \alpha_{d}
\end{aligned}
\end{equation}
This allows the network to selectively enhance or suppress specific border prompts based on the semantic characteristics of the image, resulting in a visual prompt that is better adapted to the current input.
Finally, the modulated border prompts are combined with a zero-padded central region to form a full-size prompt matching the input image dimensions.
We then augment the input image \(X\) with the modulated prompt and feed it into the pretrained privacy segmentation model to obtain the predicted segmentation results $\hat{M}$ .We employ cross-entropy loss $\mathcal{L}_{\text{CE}}$ between the predicted mask $\hat{M}$ and the ground truth mask $M$ to optimize the parameters of each style-prompt generator:
\begin{equation}
\mathcal{L}_{\text{CE}}(\hat{M}, M) = -\frac{1}{N}\sum_{i=1}^{N} \sum_{c=1}^{C} M_{i,c} \log(\hat{M}_{i,c})
\label{loss}
\end{equation}
where $N$ is the number of pixels and $C$ is the number of semantic classes. After training, each generator $G_i$ acquires effective prompting capabilities specifically tailored to its corresponding style.

\begin{figure}[t]
    \centering
    \includegraphics[width=0.3\textwidth]{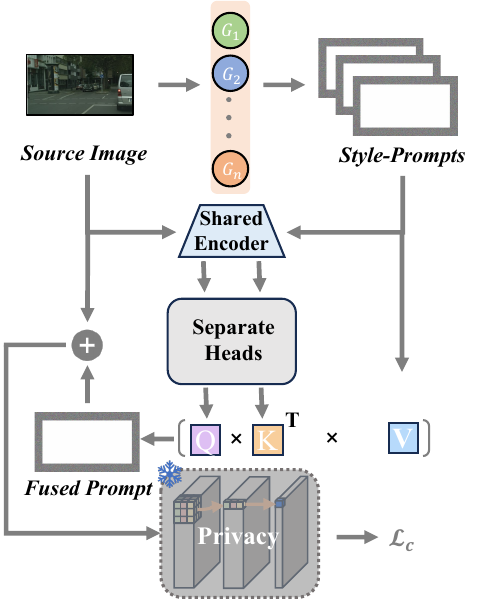}
    \caption{The overview of APF.}
    \label{fig:APF}
    \vspace{-1.0em}
\end{figure}

\subsection{Adaptive Prompt Fusion}
Although the SPG phase provides generators capable of producing diverse style-specific prompts, it remains uncertain which style prompt is most suitable when performing inference on target domain images. Moreover, a single image may exhibit characteristics of multiple styles, with varying degrees of emphasis. In this context, enabling the model to automatically select and combine the most relevant style-prompts becomes crucial.

To address this, we propose the Adaptive Prompt Fusion (APF) as shown in Figure \ref{fig:APF}, which employs a cross-attention mechanism to adaptively fuse multiple style prompts based on the input image. The fused prompt is then integrated with the image to achieve more accurate segmentation results.

Given an input image $\mathbf{X} \in \mathbb{R}^{B \times C \times H \times W}$, we first generate the corresponding style-prompts using the \(n\) style-prompt generators obtained from the SPG phase. Since the numerical scales of different prompts vary, we perform L2 norm normalization on each style-prompt. This normalization helps ensure a more balanced contribution from each prompt during the subsequent fusion process.
For each of the $n$ style-prompt generators, we generate corresponding prompts:
\begin{equation}
\mathbf{P}_i = \frac{G_i(\mathbf{X})}{||G_i(\mathbf{X})||_2}\in \mathbb{R}^{B \times C \times H \times W}, \quad i \in \{1,2,\ldots,n\}
\end{equation}

\begin{table*}[ht]
\centering
\caption{Performance comparison between our method and existing DGSS methods under settings: i) G $\rightarrow$ \{C,B,M,S\}, ii) C $\rightarrow$ \{B,M,G,S\}, iii) S $\rightarrow$ \{C,B,M,G\}. \textbf{Bold numbers} indicate the best performance among privacy methods, and \underline{underlined numbers} denotes the best among the remaining methods. }
\resizebox{1.0\linewidth}{!}{
\begin{tabular}{l@{}|c|c|cccc|c|cccc|c|cccc|c}
\toprule
\multirow{2}{*}{\textbf{Methods}}
& \multirow{2}{*}{\textbf{Privacy}}
& \multicolumn{1}{c|}{\textbf{\#Trainable}}
& \multicolumn{5}{c|}{\textbf{Trained on GTAV (G)}}
& \multicolumn{5}{c|}{\textbf{Trained on Cityscapes (C)}}
& \multicolumn{5}{c}{\textbf{Trained on SYNTHIA (S)}}
\\ \cline{4-8} \cline{9-13} \cline{14-18}
& & \textbf{Parameters} & $\rightarrow$ C & $\rightarrow$ B & $\rightarrow$ M & $\rightarrow$ S & Avg.
& $\rightarrow$ B & $\rightarrow$ M & $\rightarrow$ G & $\rightarrow$ S & Avg. & $\rightarrow$ C & $\rightarrow$ B & $\rightarrow$ M & $\rightarrow$ G & Avg. \\
\midrule
Baseline & \checkmark & 0.01M & 42.46& 35.63& 36.68& 29.86& 36.16& 40.30 & 40.06 & 38.82 & 31.05 & 37.56 & 37.07 & 30.57 & 33.41 & 35.71  & 34.19 \\

Full Fine-Tuning & & 84.61M & 45.33 & 40.29 & 39.44 & 31.50 & 39.14 & 44.44 & 43.67 & 43.02 & \underline{33.15} & 41.07 & \underline{40.01} & 33.29 & 33.52 & \underline{37.11} & \underline{35.98}  \\


IBN-Net \cite{IBN} & & 25.56M & 33.85 & 32.30 & 37.75 & 27.90 & 32.95 & 48.56 & 57.04 & 45.06 & 26.14 & 44.20 & 32.04 & 30.57 & 32.16 & 26.90 & 30.42 \\
IW \cite{IW} & & 25.56M & 29.91 & 27.48 & 29.71 & 27.61 & 28.68 & 48.49 & 55.82 & 44.87 & 26.10 & 43.82 & 28.16 & 27.12 & 26.31 & 26.51 & 27.03 \\
DRPC \cite{DRPC} & & 25.56M & 37.42 & 32.14 & 34.12 & 28.06 & 32.94 & 49.86 & 56.34 & 45.62 & 26.58 & 44.60 & 35.65 & 31.53 & 32.74 & 28.75 & 32.17 \\
ISW \cite{ISW} & & 25.56M & 36.58 & 35.20 & 40.33 & 28.30 & 35.10 & 50.73 & 58.64 & 45.00 & 26.20 & 45.14 & 35.83 & 31.62 & 30.84 & 27.68 & 31.49 \\
GTR \cite{GTR} & & 25.56M & 37.53 & 33.75 & 34.52 & 28.17 & 33.49 & 50.75 & 57.16 & 45.79 & 26.47 & 45.04 & 36.84 & 32.02 & 32.89 & 28.02 & 32.44 \\
SAW \cite{SAW} & & 25.56M & 39.75 & 37.34 & \underline{41.86} & 30.79 & 37.44 & \underline{52.95} & \underline{59.81} & 47.28 & 28.32 & \underline{47.09} & 38.92 & \underline{35.24} & \underline{34.52} & 29.16 & 34.46 \\

A2XP \cite{A2XP} & \checkmark & 1.21M & 35.71 & 28.46 & 31.22 & 22.62 & 29.50 & 32.40 & 34.68 & 31.01 & 25.74 & 30.96 & 29.79 & 25.65 & 27.22 & 26.94 & 27.40 \\


FAMix \cite{fahes2024simple} &    & 31.22M & \underline{47.16} & \underline{45.50} & 35.13 & \underline{52.17} & \underline{44.99} & 51.36 & 45.76 & \underline{57.20} & 32.19 & 46.63 & 24.57 & 29.05 & 27.92 & 24.50 & 26.51 \\ 

\rowcolor{orange!8} 
\textbf{SAGE (Ours)} & \checkmark & 1.53M & \textbf{51.38} & \textbf{39.35} & \textbf{44.12} & \textbf{33.50} & \textbf{42.09} & \textbf{45.85} & \textbf{50.27} & \textbf{47.07} & \textbf{32.40} & \textbf{43.90} & \textbf{40.87} & \textbf{35.91} & \textbf{36.29} & \textbf{37.26} & \textbf{37.58} \\
\bottomrule
\end{tabular}}
\label{tab:dgss_comparison_g}
\end{table*}

We draw inspiration from the cross-attention mechanism to obtain the attention allocation of the input image to different prompts. To ensure that the input image and style-prompts can be projected into the same feature space, we use a pre-trained shared encoder to extract their features separately. Subsequently, two separate linear heads are employed to allow the model to learn specialized representations for both, optimizing their respective functional characteristics.
By treating the \(n\) style-prompts as the values (V), their features as the keys (K), and the input image's features as the query (Q), we leverage the attention mechanism to compute the attention score for each style-prompt. 
\begin{equation}
    \mathbf{\mathbf{A}}_i = W_x(\mathcal{E}(\mathbf{X}))W_p(\mathcal{E}(\mathbf{P}_i))^T
\end{equation}
where \(\mathcal{E}\) is the shared encoder, \(W_x\) and \(W_p\) are the linear heads specifically designed for the input and prompts, respectively.
To enable the model to focus on the most relevant prompts, we apply a softmax function to the attention scores, converting them into attention weights for selecting each prompt. Furthermore, to prevent the weights from being too close to 1 and to avoid the model over-confidently favoring certain prompts, these weights are further compressed using a tanh function.
We will then perform a weighted summation of these \(n\) prompts based on $\mathbf{A}_i$ to obtain the final prompt.

\begin{equation}
    \mathbf{P}_\textbf{fused} =\sum_{i=1}^n \tanh\left(\frac{e^{\mathbf{A}_i}}{\sum_{j=1}^{n} e^{a_j}}\right)\mathbf{P}_i
\end{equation}

After that, \(p_{\text{fused}}\) is appended to the input image and fed into the pre-trained model to obtain the predicted segmentation results, which are then used to compute the cross-entropy with the ground truth segmentation in the same manner as Equation \ref{loss}. However, we only optimize the separate heads based on the loss function in this phase since the shared encoder has already been pre-trained, which ensures our model remains lightweight.

The adaptive fusion mechanism allows the model to leverage multiple style-prompts simultaneously, weighting them according to their relevance to the image's styles, thereby achieving more effective segmentation across target domains. Once the APF phase is complete, we can perform inference on unseen target domains under privacy constraints. More details can be found in the Appendix \ref{theoinf}. 
\section{Experiments}
\label{res}
In this section, we describe the implementation details of our approach, the setup for comparison with existing DGSS methods, and the ablation studies conducted to further validate the effectiveness of our method across five datasets: GTAV \cite{richter2016playing} (G), SYNTHIA \cite{synthia} (S), Cityscapes \cite{cityscape} (C), BDD-100K \cite{bdd100k} (B), and Mapillary \cite{mapillary} (M) (Table \ref{tab:dataset_summary}). The details of each dataset are provided in the Appendix \ref{dataset}.

\begin{table}[h]
\centering
\small
\caption{Summary of the five datasets used in our experiments.}
\label{tab:dataset_summary}
\resizebox{1.0\linewidth}{!}{
\begin{tabular}{lccccc}
\toprule
\textbf{Dataset} & \textbf{Type} & \textbf{Resolution} & \textbf{Train Images} & \textbf{Val Images}\\
\midrule
GTAV       & Synthetic & 1914×1052 & 12,403 & 6,382  \\
SYNTHIA    & Synthetic & 1280×760  & 6,580  & 2,820  \\
Cityscapes & Real      & 2048×1024 & 2,975  & 500    \\
BDD-100K   & Real      & 1280×720  & 7,000  & 1,000  \\
Mapillary  & Real      & Varied    & 18,000 & 2,000  \\
\bottomrule
\end{tabular}}
\vspace{-1em}
\end{table}

\subsection{Implementation details}
\label{set}
We use SegFormer-B5 \cite{xie2021segformer}, pretrained on ADE20K \cite{zhou2017scene}, as the privacy segmentation model.
In the SPG phase, each style-prompt generator is trained for 10,000 iterations with a batch size of 2. The prompt template for each generator is initialized with a padding size of 30. We adopt stochastic gradient descent (SGD) with momentum (0.9) as the optimizer and set the initial learning rate to 1e-4.
In the APF phase, we use an ImageNet \cite{deng2009imagenet}-pretrained ResNet18 as the shared encoder, along with two linear layers as task-specific heads. The model is trained for 40,000 iterations with a batch size of 2. Optimization is performed using the AdamW optimizer with an initial learning rate of 1e-4. All experiments are conducted on an NVIDIA RTX 4090 GPU.

\subsection{Comparison with state-of-the-art}
\label{com}

We compare our method with several state-of-the-art DGSS methods including IBN-Net \cite{IBN}, IW \cite{IW}, DRPC \cite{DRPC}, ISW \cite{ISW}, GTR \cite{GTR}, SAW \cite{SAW}, FAMix \cite{fahes2024simple} as well as A2XP \cite{A2XP}, a domain generalization method originally designed for image classification in the privacy setting. Full Fine-Tuning means training directly on the source domain and fine-tuning the backbone of the privacy model.
To evaluate the generalization performance on unseen domains, we conduct experiments under three cross-domain setups:
i) G $\rightarrow$ \{C,B,M,S\}, ii) C $\rightarrow$ \{B,M,G,S\}, iii) S $\rightarrow$ \{C,B,M,G\}.
We use mIoU(\%) as the evaluation metric.
All reported results are cited directly from previous works \cite{ISW,IBN,IW,SAW}.



Table \ref{tab:dgss_comparison_g} summarizes the performance of SAGE and several DGSS baselines under three different setups. Compared with existing privacy-preserving approaches, SAGE demonstrates clear advantages across all transfer directions and achieves results that are comparable to or even surpass those of non-privacy methods. This highlights its strong balance between domain generalization capability and parameter efficiency.
When trained on GTAV, SAGE reaches an average mIoU of 42.09\%, outperforming the privacy baseline by more than 2 percentage points. Particularly strong improvements are observed on Cityscapes (51.38\%) and Mapillary (44.12\%), indicating that SAGE effectively preserves semantic consistency across synthetic-to-real transitions. The structured adaptive design enables the model to remain robust under substantial scene variations and appearance gaps. When Cityscapes serves as the source domain, SAGE again achieves the best performance among privacy methods with an average mIoU of 43.90\%, showing remarkable accuracy on BDD-100K (45.85\%) and GTAV (47.07\%). Despite using only 1.53 M trainable parameters, SAGE surpasses many normalization- or style-based adaptation methods such as ISW and SAW, and even approaches or exceeds several full-parameter fine-tuning models, demonstrating its parameter efficiency and adaptability under privacy constraints. Under the SYNTHIA-to-real scenario, SAGE maintains strong robustness with an average mIoU of 37.58\%, the highest among all privacy-preserving methods. Even with the significant domain gap between synthetic and real data, the model performs competitively on Cityscapes (40.87\%) and GTAV (37.26\%), revealing its ability to learn transferable and generalizable representations from low-fidelity synthetic inputs. Overall, across all three domain configurations, SAGE achieves 3–5\% higher average accuracy than the baseline and about 12–14\% improvement over A2XP, while requiring dramatically fewer trainable parameters. These results clearly validate the effectiveness of adaptive feature fusion and sample-sensitive modulation mechanisms, which together yield robust cross-domain generalization and privacy-preserving performance across diverse urban visual environments.

\subsection{Ablation studies}
\label{ab}

\begin{figure*}[t]
  \centering
  \includegraphics[width=\textwidth]{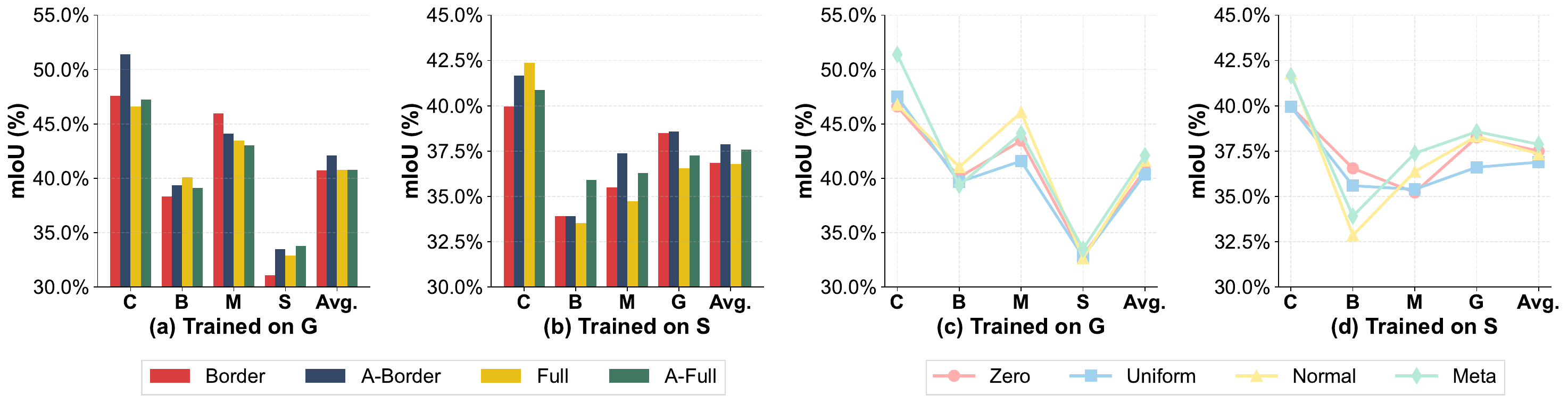}
  \caption{Ablation study on prompt design. (a) and (b) show the mIoU performance comparison between different prompt generator types on the G $\rightarrow$ \{C, B, M, S\} and S $\rightarrow$ \{C, B, M, G\} tasks, respectively. (c) and (d) evaluate the impact of different prompt template initialization strategies.}
  \label{fig:prompt}
\end{figure*}
\begin{figure*}[t]
  \centering
  \vspace{-0.5em}
  \includegraphics[width=\textwidth]{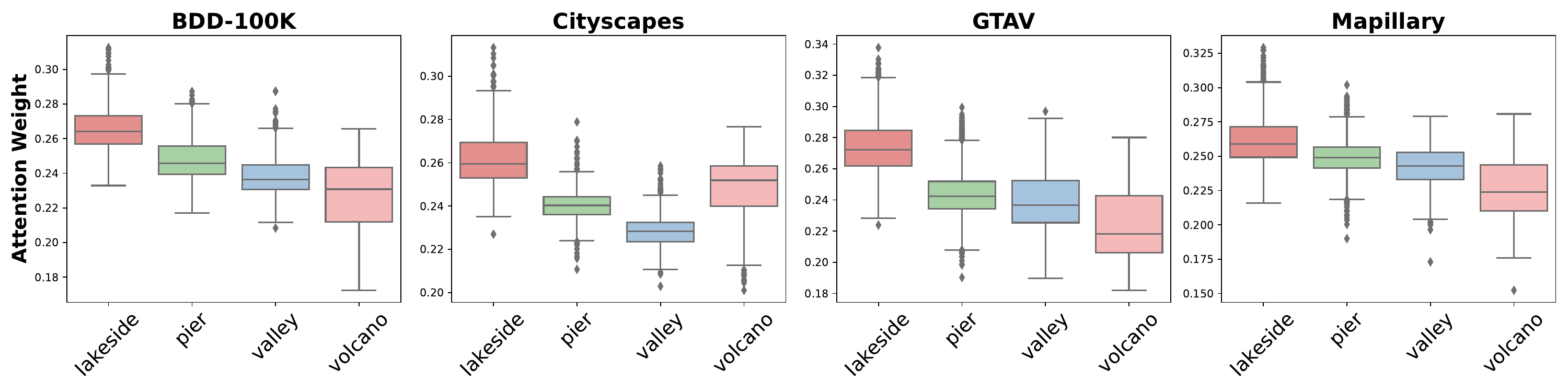}
  \vspace{-1em}
  \caption{Attention weight distribution over four style-prompts (lakeside, pier, valley, volcano) for images from different target domains (BDD-100K, Cityscapes, GTAV, Mapillary). The results indicate that each domain exhibits distinct style preferences. }
  \label{fig:attn}
  \vspace{-0.5em}
\end{figure*}

\textbf{On Different Generator types.} Figure \ref{fig:prompt} (a) and (b) illustrates the impact of different types of prompt generators on mIoU performance. "A" denotes "adaptive", indicating that a modulation network is used to dynamically reweight the prompt template based on the input. Border and Full refer to fixed prompt applied to all images, the former along the image boundaries and the latter across the full image area. 
Both A-Border and A-Full outperform their fixed counterparts, demonstrating that even for images of the same style, the effectiveness of a given prompt can vary. This highlights the necessity of the modulation network, which adaptively adjusts the prompt to better suit individual samples. Furthermore, A-Border generally yields better results than A-Full, as it provides auxiliary information while minimizing interference with the original image content. These results confirm that adaptive reweighting are critical to effective prompt-based domain generalization.


\textbf{On Different Initialization Strategies.} Figure~\ref{fig:prompt} (c) and (d) present results for various initialization strategies of prompt templates. Meta refers to a meta-initialization strategy, where each prompt generator is first pretrained with a small number of iterations on each style-specific subset of the source domain to acquire basic prompting ability. It is then further fine-tuned in the SPG phase to focus on a specific style.
Among zero initialization, uniform distribution, normal distribution, and Meta, the Meta strategy shows a clear advantage. This is because the meta-initialization allows the prompt to start from a semantically meaningful state, enabling faster convergence and better alignment with style-specific features during the SPG phase. 

\begin{table}[t]
    \centering
    \small
    \caption{Ablation study on \textit{Cityscapes} for different optimizations within the APF phase.}
    \begin{tabular}{ccc|c}
    \toprule
    \multicolumn{3}{c|}{\textbf{Attention Optimization}} & \multirow{2}{*}{\textbf{Avg. mIoU}} \\ \cline{1-3}
    \textit{Prompt Normalization} &  \textit{Softmax}  & \textit{tanh} &  \\
    \midrule
     &  &  & 40.63 \\
    \checkmark &  &  & 41.85 \\
     & \checkmark &  & 40.49 \\
     &  & \checkmark & 41.59 \\
    \checkmark & \checkmark &  & 42.17 \\
     & \checkmark & \checkmark & 41.46 \\
    \checkmark &  & \checkmark & 41.16 \\
    \checkmark & \checkmark & \checkmark & \textbf{43.90} \\
    \bottomrule
    \end{tabular}
    \label{tab:fusion}
    \vspace{-1.5em}
\end{table}

\begin{figure*}[t]
  \centering
  \includegraphics[width=\textwidth]{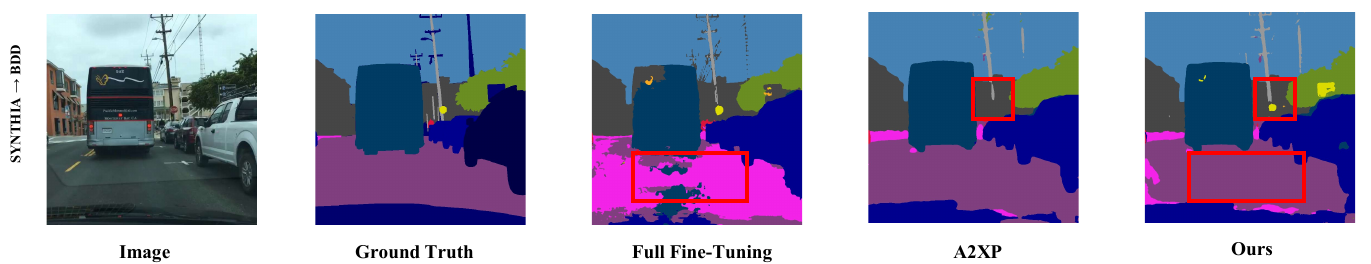}
  \caption{Visual segmentation results on B under S $\rightarrow$ \{C, B, M, G\} setting. The proposed SAGE is compared with Full Fine-Tuning and A2XP. Compared to alternative methods, SAGE yields less noise and demonstrates a better ability to capture details.}
  \vspace{-0.5em}
  \label{fig:visual}
\end{figure*}

\textbf{On Attention Optimization.} Table~\ref{tab:fusion} shows the effect of different component combinations within the APF phase on the average mIoU when trained on Cityscapes (C). Prompt normalization (PN) performs L2 normalization on each style-prompt. Enabling PN improves the average mIoU from 40.63 to 41.85, indicating that normalizing prompts to a similar scale ensures a more balanced contribution from each style during the fusion process.
Building upon this, incorporating a softmax operation converts attention scores into a probability distribution, allowing the model to better prioritize the most relevant prompts. This leads to a further improvement from 41.85 to 42.17 in average mIoU. However, to prevent the model from becoming overly reliant on a single prompt where one weight approaches 1, applying a tanh function to compress the attention weights yields the best result, achieving a 1.73\% gain, and a total improvement of 3.27\% over the baseline with no components.
Other combinations are slightly less effective. Notably, using softmax alone without normalization leads to worse performance, likely due to the unbalanced scaling of the input prompts, which skews the weighted summation and hinders generalization.


\subsection{Quantitative analysis of style attention}



As shown in Figure \ref{fig:attn}, to further validate the existence of diverse style characteristics across datasets, we analyze the attention weights assigned to different style prompts under the S $\rightarrow$ {C, B, M, G} setting. Four representative style references from ImageNet including lakeside, pier, valley, and volcano are selected to capture distinct visual styles. The results reveal that samples from different domains exhibit clear preferences for specific styles. For example, Cityscapes images consistently assign higher weights to the volcano style prompt, likely due to their typical daylight capture conditions, which visually resemble the vivid tones of the volcano style. In contrast, the volcano style receives much lower weights in other domains. Interestingly, the lakeside style, characterized by elements such as sky and horizon, receives the highest attention weights across all domains, which is reasonable given that these visual features are commonly present in all datasets. Meanwhile, the pier and valley styles show varying contributions depending on the domain, reflecting differences in urban and rural scene composition. These observations confirm the presence of diverse visual styles and demonstrate how our SAGE framework effectively adapts to this challenge by dynamically combining multiple style-prompts.

\subsection{Visualization analysis}

Figure \ref{fig:visual} illustrates the segmentation results of Full Fine-Tuning, A2XP, and our method when trained on SYNTHIA and evaluated on BDD-100K.
Full Fine-Tuning, which simply adapts a pre-trained model to the source domain through naive fine-tuning without any generalization mechanisms, compromises the model’s inherent generalization ability. As a result, it performs well only on the source domain but generates substantial noise when inferring on unseen domains, leading to fragmented and irregular object boundaries in the segmentation results.
In contrast, A2XP employs a style-fixed prompt to guide the privacy model. However, this rigidity causes the method to miss fine details when segmenting images with styles different from the fixed one. For instance, the two traffic signs in the figure are not recognized in its results.
Our method, on the other hand, introduces an adaptive prompt that dynamically aligns with the style of the input image while keeping the pre-trained model’s backbone frozen, which enables our approach to reduce noise while preserving fine details. As a result, our segmentation results maintain complete object boundaries and successfully capture subtle elements such as the inconspicuous traffic signs in the scene. 
More visual results can be found in the Appendix \ref{more_visual}.
\section{Conclusion}
\label{conc}

In this work, we formalize the privacy-constrained generalization as a critical deployment-aware challenge in DGSS, where frozen models necessitate input-level adaptation, addressed here for the first time.
We propose the SAGE framework, which leverages the strong representation capabilities of pretrained models and external visual prompts. Since unseen target domains often exhibit styles absent from the source domain, we first augment the source data via style transfer with an auxiliary dataset. We then train a set of style-prompt generators, each specialized for a particular style and capable of producing prompts conditioned on the input image. These prompts are then fused using a cross-attention mechanism based on the input query, generating an adaptive style-aware prompt that is integrated with the image to guide the black-box model toward better segmentation. Extensive experiments across three cross-domain settings show that SAGE achieves competitive or superior performance compared to state-of-the-art methods under privacy constraints and consistently outperforms full fine-tuning baselines. 

\section{Acknowledgements}
This work was supported in part by the National Natural Science Foundation of China under Grant T2125006 and Grant 42401415; in part by Shenzhen Science and Technology Program under Grant KCXFZ20240903093759004 and Grant KJZD20230923115106012; in part by Guangdong Science \& Technology Program under Grant 2025B0101080001; in part by Tsinghua SIGS KA Cooperation Fund.
{
    \small
    \bibliographystyle{ieeenat_fullname}
    \bibliography{main}
}

\clearpage
\setcounter{page}{1}
\maketitlesupplementary



SAGE enables domain generalization for target models without requiring access to the internal parameters, architecture of the pre-trained models, or large-scale training datasets. This allows the target models to be widely applied across various scenarios. Our research holds significant promise for real-world applications such as autonomous driving, medical image analysis, and remote sensing, where sensitive data and proprietary models are often involved. By achieving strong domain generalization without revealing model internals or requiring private data, our method addresses the increasing demand for privacy-preserving deep learning. SAGE can facilitate the deployment of robust vision systems in areas like healthcare and urban planning, supporting more informed decision-making while safeguarding user privacy. Furthermore, industries relying on third-party models accessed via APIs can benefit from improved performance on unseen domains without compromising security.

The supplementary is organized as follows:

\begin{itemize}[leftmargin=1.5em] 
    \item In Sec.~\ref{style_transfer}, we provide a comprehensive description of the style transfer technique, including implementation details and representative examples of translation.
    \item Sec.~\ref{theoretical} presents the theoretical foundations of the proposed SAGE framework, covering style-prompt generation, adaptive prompt fusion, and the inference process.
    \item Sec.~\ref{ex_app} describes the experimental setup in detail, including dataset specifications, model configurations, training hyperparameters, and computational resources.
    \item Sec.~\ref{more_exp} reports additional experimental investigations, including efficiency comparisons, ablation studies, and extended qualitative segmentation results.
    \item Sec.~\ref{limitation} discusses the limitations of our approach and outlines potential directions for future research.
\end{itemize}


\section{Style transfer technique}
\label{style_transfer}

\subsection{Implementation details}
We adopt a CycleGAN-based \cite{zhu2017unpaired} approach to perform style augmentation on the source domain. CycleGAN offers powerful unpaired image-to-image translation capabilities, making it well-suited for our domain generalization task. The model consists of two generators, $G_A$and $G_B$, which learn mappings from the source domain to the target style and vice versa. Each generator is built on a ResNet architecture with 9 residual blocks, allowing effective style transformation while preserving image content.

To improve the realism of generated images, we employ a PatchGAN discriminator that distinguishes between real and translated images. The model is optimized using the original CycleGAN objective, which combines multiple loss terms including adversarial loss, cycle-consistency loss, and identity loss. Training proceeds in two alternating steps: first, updating the generators to fool the discriminators; then, updating the discriminators to better distinguish real from generated images.

We use the Adam optimizer with an initial learning rate of 0.0002 and apply a linear decay schedule. To enhance training stability, we incorporate a history buffer that stores previously generated images for training the discriminator. Model checkpoints are saved every 50 epochs, and intermediate results are visualized to monitor training progress. After training, the style transformation for any input image is performed via a simple forward pass through the generator.

This CycleGAN-based method enables effective style translation on the source domain and provides a strong data augmentation strategy to support our domain generalization framework.

\begin{figure*}[htb]
  \centering
  \includegraphics[width=\textwidth]{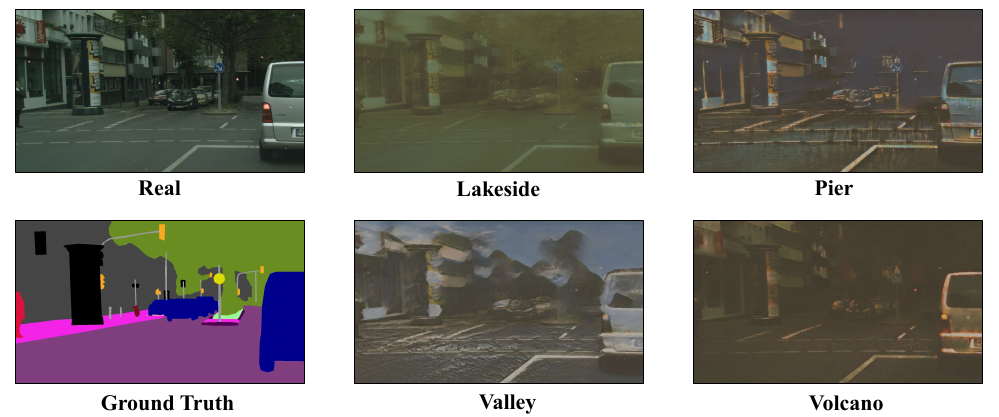}
  \vspace{-1.5em}
  \caption{Examples of CycleGAN-based style augmentation.}
  \label{fig:styles}
\end{figure*}

\subsection{Examples of translation}
 Examples of our style augmentation results are illustrated in Figure \ref{fig:styles}. We select four representative style categories from ImageNet including lakeside, pier, valley, and volcano to simulate distinct environmental appearances commonly seen in real-world driving scenarios. For example: lakeside is used to simulate rainy scenes due to its frequent inclusion of overcast skies, reflections, and blurred contours, mimicking the reduced visibility and color saturation typical of rainy conditions. Pier, often characterized by structured man-made elements like docks and buildings, is used to replicate architectural styles and urban textures. Valley, dominated by open natural landscapes and vegetation, captures the essence of rural or countryside environments. Volcano, with its hazy, low-contrast tones and dramatic lighting, effectively simulates foggy or low-visibility scenes. By performing style transfer using these categories, we can generate stylized versions of the source domain data that mimic various environmental conditions.

\section{Theoretical details of SAGE}
\label{theoretical}

\subsection{Style-prompt generation}
\label{theoSPG}
Figure \ref{fig:SPG} provides an overview of the SPG phase architecture. Considering that the target domain may include diverse real-world scenes whose visual styles deviate substantially from the source domain, it becomes crucial to generate style-aware prompts that can generalize effectively across such variations. To tackle this challenge, we randomly select $n$ categories from ImageNet \cite{deng2009imagenet}, which serves as an auxiliary style dataset. These selected categories provide diverse style references. We then employ an CycleGAN translation approach \cite{zhu2017unpaired} to render source images in the visual style of each reference category. 

Nevertheless, visual similarity alone does not guarantee semantic consistency across scenes, prompt relevance may still vary due to object composition and contextual layout. Thus, relying on a single fixed prompt per style reference would be insufficient. To address this limitation, we introduce a dynamic style-prompt generator, which produces content-adaptive prompts tailored to each input image.

Each generator $G_i$ is associated with a particular style $i \in \{1,2,\ldots,n\}$ and contains a learnable prompt template $\mathcal{T}_i = \{P_t, P_b, P_l, P_r\}$ with learnable weights only located at the four borders (top, bottom, left, and right), while the central region is explicitly zeroed out. This design captures style-related prior knowledge in a spatially constrained form, focusing the modulation on semantically informative boundaries.

To adapt the template to image content, a lightweight modulation network $\mathcal{M}$ is used to compute attention weights that adjust each border prompt. This modulation network is composed of stacked residual blocks, termed ModulatorBlocks, followed by global feature aggregation. The forward function of each ModulatorBlock can be written as:

\begin{equation}
    f_\text{block}(\mathbf{X}; \theta_i) = \mathcal{F}(\mathbf{X}; \theta_i) + \mathcal{S}(\mathbf{X}; \phi_i)
\end{equation}

where $\mathcal{F}(\cdot)$ denotes the main convolutional path, while $\mathcal{S}(\cdot)$ is the shortcut pathway, and $\theta_i$, $\phi_i$ are their respective learnable parameters. Specifically, each main path takes the form:

\begin{equation}
    \mathcal{F}(\mathbf{X}; \theta_i) = \text{BN}_2(\text{Conv}_2(\sigma(\text{BN}_1(\text{Conv}_1(\mathbf{X})))))
\end{equation}

where $\sigma$ denotes the ReLU activation function.
The network processes the input through three stages with increasing channel dimensions and progressively lower spatial resolutions:

\begin{align}
    \mathbf{X}_1 &= f_\text{block1}(\mathbf{X}; \theta_1) \in \mathbb{R}^{B \times D_1 \times \frac{H}{2} \times \frac{W}{2}} \\
    \mathbf{X}_2 &= f_\text{block2}(\mathbf{X}_1; \theta_2) \in \mathbb{R}^{B \times D_2 \times \frac{H}{4} \times \frac{W}{4}} \\
    \mathbf{X}_3 &= f_\text{block3}(\mathbf{X}_2; \theta_3) \in \mathbb{R}^{B \times D_3 \times \frac{H}{8} \times \frac{W}{8}}
\end{align}

where $D_1 = d$, $D_2 = 2d$, and $D_3 = 4d$. These representations encode increasingly abstract semantic information.
The final stage involves projecting $\mathbf{X}_3$ to a $4C$-channel feature map, then applying adaptive average pooling to yield a compact global descriptor:

\begin{align}
    \mathbf{X}_4 &= \text{Conv}_{1\times1}(\mathbf{X}_3) \in \mathbb{R}^{B \times 4C \times \frac{H}{8} \times \frac{W}{8}}\\
    \mathcal{M}(\mathbf{X}) &= \text{AdaptiveAvgPool}(\mathbf{X}_4) \in \mathbb{R}^{B \times 4C \times 1 \times 1}
\end{align}

Next, the output is reshaped into modulation coefficients for each border region:

\begin{equation}
    \alpha = \text{reshape}(\mathcal{M}(\mathbf{X})) \in \mathbb{R}^{B \times 4 \times C \times 1 \times 1}
\end{equation}

where $\alpha = [\alpha_p, \alpha_b, \alpha_l, \alpha_r]$ contains the modulation coefficients for each border.
The original semantic border $P_t, P_b, P_l, P_r \in \mathbb{R}^{B \times C \times H_p \times W_p}$ are modulated as:
\begin{equation}
\begin{aligned}
    P'_t = P_t \odot \alpha_{:,0,:,:,:}, \\
    P'_b = P_b \odot \alpha_{:,1,:,:,:}, \\
    P'_l = P_l \odot \alpha_{:,2,:,:,:}, \\
    P'_r = P_r \odot \alpha_{:,3,:,:,:}
\end{aligned}
\end{equation}
where $\odot$ denotes element-wise multiplication with appropriate broadcasting.
Let $Z \in \mathbb{R}^{B \times C \times H_c \times W_c}$ be a zero tensor representing the center region. The final adaptive border prompt is assembled via spatial concatenation:

\begin{equation}
    P_{i} = \begin{bmatrix} 
    P'_t \\
    P'_l, Z, P'_r \\
    P'_b
    \end{bmatrix}
\end{equation}

This formulation enables the network to generate style-prompts by adaptively modulating each border region based on the semantic content of the input image.

We then augment the input image $\mathbf{X}$ with style-prompts $P_i$ before feeding it to a pre-trained black-box segmentation model $\Phi$. The augmented input can be formulated as:

\begin{equation}
\mathbf{X}' = \mathbf{X} \oplus G_i(\mathbf{X})
\end{equation}

where $\oplus$ denotes the prompt attachment operation and $G_i$ is the style-prompt generator corresponding to style $i$. The black-box model then produces a predicted segmentation mask:

\begin{equation}
\hat{M} = \Phi(\mathbf{X}')
\end{equation}

We employ cross-entropy loss $\mathcal{L}_{\text{CE}}$ between the predicted mask $\hat{M}$ and the ground truth mask $M$ to optimize the parameters $\theta_s$ of each style-prompt generator:

\begin{equation}
\mathcal{L}_{\text{CE}}(\hat{M}, M) = -\frac{1}{N}\sum_{i=1}^{N} \sum_{c=1}^{C} M_{i,c} \log(\hat{M}_{i,c})
\label{app:loss}
\end{equation}

where $N$ denotes the total number of pixels, and $C$ is the number of semantic categories. After training, each generator $G_i$ is capable of producing effective prompts customized to the specific style it represents.

\subsection{Adaptive Prompt Fusion}
\label{theoAPF}
In the second stage of our framework, we aim to dynamically determine which style prompts are most appropriate for a given target domain image. Despite the availability of diverse style-specific prompts from SPG, real-world images often exhibit a mixture of multiple styles. Therefore, a mechanism that adaptively weighs and combines these prompts is essential.

We introduce the Adaptive Prompt Fusion (APF), which utilizes a cross-attention strategy to compute the relevance between the input image and each candidate prompt. Based on these computed weights, it performs a prompt fusion to enhance segmentation performance across various target styles.
Given an input image $\mathbf{X} \in \mathbb{R}^{B \times C \times H \times W}$, we generate a collection of style prompts using $n$ pre-trained generators $G_1, \ldots, G_n$:

\begin{equation}
\mathbf{P}_i = G_i(\mathbf{X}), \quad i \in \{1,2,\ldots,n\}
\end{equation}

These are then stacked along a new dimension to form a 5D tensor:

\begin{equation}
\mathbf{P} = [\mathbf{P}_1, \mathbf{P}_2, \ldots, \mathbf{P}_K]\in \mathbb{R}^{B \times N \times C \times H \times W}
\end{equation}

We then apply L2 normalization over spatial dimensions to reduce scale discrepancies among prompts:

\begin{equation}
\mathbf{P}_{\text{norm}} = \frac{\mathbf{P}}{||\mathbf{P}||_2}
\end{equation}

To evaluate how relevant each prompt is to the input image, we employ a pre-trained shared encoder $\mathcal{E}$ to extract feature representations. The input image is encoded as:

\begin{align}
\mathbf{f}_X &= \mathcal{E}(\mathbf{X}) \in \mathbb{R}^{B \times D}
\end{align}

which is projected into a query embedding space using a learnable projection head:

\begin{align}
\mathbf{e}_X &= W_x(\mathbf{f}_X) \in \mathbb{R}^{B \times d}
\end{align}

For the prompts, we first flatten the $K$ variants per image into a batch format:
\begin{align}
\mathbf{P}_{\text{flat}} \in \mathbb{R}^{(B \cdot N) \times C \times H \times W}
\end{align}

Then encode and project them:
\begin{align}
\mathbf{f}_P &= \mathcal{E}(\mathbf{P}_{\text{flat}}) \in \mathbb{R}^{(B \cdot N) \times D} \\
\mathbf{e}_P &= W_p(\mathbf{f}_P) \in \mathbb{R}^{(B \cdot N) \times d}
\end{align}
We reshape $\mathbf{e}_P$ back to group the $n$ prompts per sample:
\begin{align}
\mathbf{E}_P &= \text{reshape}(\mathbf{e}_P) \in \mathbb{R}^{B \times N \times d}
\end{align}

Finally, cross-attention scores are computed by dot product between the input query and prompt keys:
\begin{equation}
\mathbf{\alpha}_{\text{attn}} = \mathbf{e}_X \cdot \mathbf{E}_P^T \in \mathbb{R}^{B \times N}
\end{equation}

To derive the final attention weights for prompt fusion, we first normalize the attention scores using a softmax function:
\begin{equation}
\mathbf{\alpha} = \text{softmax}(\mathbf{\alpha}_{\text{attn}}, \text{dim}=1)
\end{equation}
To avoid overconfidence in individual prompts and ensure a smoother distribution, the resulting weights are further compressed via a hyperbolic tangent:
\begin{equation}
\mathbf{\alpha} = \tanh(\mathbf{\alpha})
\end{equation}

The attention weights are reshaped to $\mathbf{\alpha} \in \mathbb{R}^{B \times N \times 1 \times 1 \times 1}$ for broadcasting. The weighted prompts are computed as:
\begin{equation}
\mathbf{P}_{\text{weighted}} = \mathbf{\alpha} \odot \mathbf{P} \in \mathbb{R}^{B \times N \times C \times H \times W}
\end{equation}
where $\odot$ represents element-wise multiplication with broadcasting.

The fused prompt is obtained by summing across the style dimension:
\begin{equation}
\mathbf{P}_{\text{fused}} = \sum_{i=1}^{N} \mathbf{P}_{\text{weighted}}[:,i,:,:,:] \in \mathbb{R}^{B \times C \times H \times W}
\end{equation}

Finally, the fused prompt is integrated with the input and fed into the black-box model to obtain the predicted results:
\begin{equation}
\mathbf{X}' = \mathbf{X} \oplus \mathbf{P}_{\text{fused}}
\end{equation}

The separate
heads $W_x$ and $W_p$ are optimized based on the loss function in the same manner as Equation \ref{app:loss}.
This adaptive fusion mechanism allows the model to leverage multiple style prompts simultaneously, weighting them according to their relevance to the input image's characteristics, thereby achieving more effective segmentation across diverse visual domains.

\subsection{Inference phase}
\label{theoinf}

\begin{algorithm}
    \caption{Inference Stage}
    \KwIn{\begin{tabular}[t]{@{}l}
    prompt generators $G_1, G_2, ..., G_n$ \\
    shared encoder $\mathcal{E}$ \\
    separate heads $W_x$ and $W_p$ \\
    input image $x$ \\
    pretrained model $M$
    \end{tabular}}
    \KwOut{Segmentation result $y$}
    \For{$i=1$ \KwTo $n$}{
      $p_i \gets G_i(x)$ \\
      \For{channel $c$ in $p_i$}{
        $p_i^c \gets \mathrm{Normalize}(p_i^c)$
      }
    }
    $V \gets [p_1; p_2; ...; p_n]$ \\
    $[f_{\text{prompt}}, f_{x}] \gets \mathcal{E}(V,x)$ \\
    $K \gets W_p(f_{\text{prompt}})$ \\
    $Q \gets W_x(f_{x})$ \\
    $A \gets Q K^\top$ \\
    $A \gets \mathrm{softmax}(A)$ \\
    $A \gets \tanh(A)$ \\
    $V' \gets A V$ \\
    $y \gets M(x + V')$ \\
    \Return $y$
    \label{infer}
\end{algorithm}

Once the SPG phase and the APF phase are completed, inference on images from an unseen target domain can be performed using the $n$ trained style-prompt generators and the separate heads. The pseudo-code for the inference process is provided in Algorithm~\ref{infer}.

Specifically, given an input image $x$, we first feed it sequentially into each of the $n$ style-prompt generators $G_i$ to obtain $n$ style-specific prompts. Each channel of these prompts is then normalized using L2 normalization to ensure consistent scale across channels.

Next, the $n$ normalized prompts are concatenated and passed, along with the image $x$, through the shared encoder $\mathcal{E}$. The resulting features are then split and fed into two separate heads to generate the key ($K$) and query ($Q$) matrices.

We compute the attention scores by multiplying $Q$ with the transpose of $K$, followed by a softmax operation and a subsequent $\tanh$ activation to produce the attention weights for each prompt. These weights are used to compute a weighted sum of the $n$ style-prompts, resulting in the final fused prompt.

Finally, this fused prompt is integrated with the input image and fed into the pre-trained black-box model to obtain the semantic segmentation prediction.

\section{Experimental details}
\label{ex_app}

\subsection{Datasets details}
\label{dataset}

We evaluate our method on five widely-used datasets for domain generalization in semantic segmentation. During training and evaluation, all images from the datasets are randomly cropped and resized to 512×512 resolution.

\textbf{GTAV(G).} GTAV is a synthetic dataset rendered using the Grand Theft Auto V game engine. It contains 24,966 simulated urban scene images with a resolution of 1914×1052, annotated with 19 semantic classes aligned with Cityscapes.

\textbf{SYNTHIA(S).} SYNTHIA is a synthetic dataset containing 9,400 images of resolution 1280×760. Following prior works, we use the SYNTHIA-RAND-CITYSCAPES subset and evaluate on 16 shared categories with Cityscapes. The dataset provides various urban scenes under different environmental conditions.

\textbf{Cityscapes(C).} This real-world dataset is collected from urban street scenes across German cities. It contains 2,975 finely annotated images for training and 500 for validation. The original resolution is 2048×1024, and annotations cover 19 semantic classes.

\textbf{BDD-100K(B).} BDD-100K includes 7,000 training and 1,000 validation images with fine pixel-level annotations across 19 categories. The dataset captures diverse driving scenes under various weather, lighting, and time-of-day conditions. The original resolution is 1280×720.

\textbf{Mapillary(M).} Mapillary provides 25,000 images collected from diverse geographic regions and devices. The images vary in resolution and are captured under a wide range of weather and scene conditions. For consistency with other datasets, we re-map its annotations to the 19 classes used in Cityscapes.

\subsection{Model and training hyperparameters}
\label{hyper}

\textbf{Architecture.}  In this study, the black-box segmentation model adopts the SegFormer architecture (B5 variant), with pretrained weights sourced from the ADE20K dataset (\texttt{nvidia/segformer-b5-finetuned-ade-640- 640}). The model structure is implemented using the HuggingFace Transformers library. We adapt the classifier layer in the decode head to align the output number of classes with the Cityscapes dataset. After passing through the backbone encoder, the input image is processed by the decode head to generate logits, which are then upsampled to the original resolution via bilinear interpolation.
The modulation network consists of three stacked \texttt{ModulatorBlock}s with channel dimensions of 32, 64, and 128, respectively, and a stride of 2 at each layer. A ResNet-18 serves as the shared encoder, initialized with ImageNet-pretrained weights. Its backbone outputs features with a dimensionality of 512, which are then projected to the embedding space (dimension 512) through two fully connected separate heads.

\textbf{Initialization.}  The prompt template is initialized using a normal distribution with a mean of 0 and a standard deviation of 0.1 before meta initialization. All convolutional layers in the modulation network are initialized with Kaiming normal initialization (with $a=0$, \texttt{mode='fan\_in'}, and \texttt{nonlinearity='leaky\_relu'}) to match the ReLU activation function. The bias terms of the convolutional layers are initialized to zero.

\textbf{Optimization and Schedule.}  During the SPG phase, we use Stochastic Gradient Descent (SGD) as the optimizer, with an initial learning rate of $1 \times 10^{-4}$ and a momentum of 0.9. A MultiStepLR scheduler is employed to decay the learning rate by a factor of 0.1 at epochs 150, 180, and 210.
In the APF phase, we switch to the AdamW optimizer with an initial learning rate of $1 \times 10^{-4}$ and $\beta$ parameters set to $(0.5,\ 0.999)$. The learning rate is scheduled using CosineAnnealingWarmRestarts, with $T_0 = 1$, $T_{mult}$ equal to the total number of epochs, and a minimum learning rate of $1 \times 10^{-5}$.

\subsection{Details of resources used}
\label{resources}

During the SPG phase, we used approximately 1 hour of GPU time per style-prompt generator on an NVIDIA RTX 4090 GPU, with around 10 GB of memory usage, utilizing the PyTorch library. In the APF phase, training took approximately 4 hours on the same GPU with about 13 GB of memory consumption, also using PyTorch. Most inferences were conducted on an RTX 4090 GPU, with computational costs that were negligible compared to the training phase.

\section{More experiments}
\label{more_exp}

\subsection{Comparison of efficiency}

\begin{table}[htbp]
\centering
\caption{Comparison of different methods in terms of parameter count, average training time per iteration, and inference time.}
\resizebox{1.0\linewidth}{!}{
\begin{tabular}{lccc}
\toprule
\multicolumn{1}{l}{\textbf{Method}} & \textbf{\#Params (M)} & \textbf{Training Time} & \textbf{Inference Time} \\
\midrule
Full Fine-Tuning & 84.61 & 371.32 & 23.26 \\
SPG Phase & 1.00 & 134.08 & - \\
APF Phase & 0.53 & 272.97 & 38.09 \\
Total & 1.53 & 407.05 & 38.09 \\
\bottomrule
\end{tabular}}
\label{tab:speed}
\end{table}

\begin{figure*}[t]
  \centering
  \includegraphics[width=\textwidth]{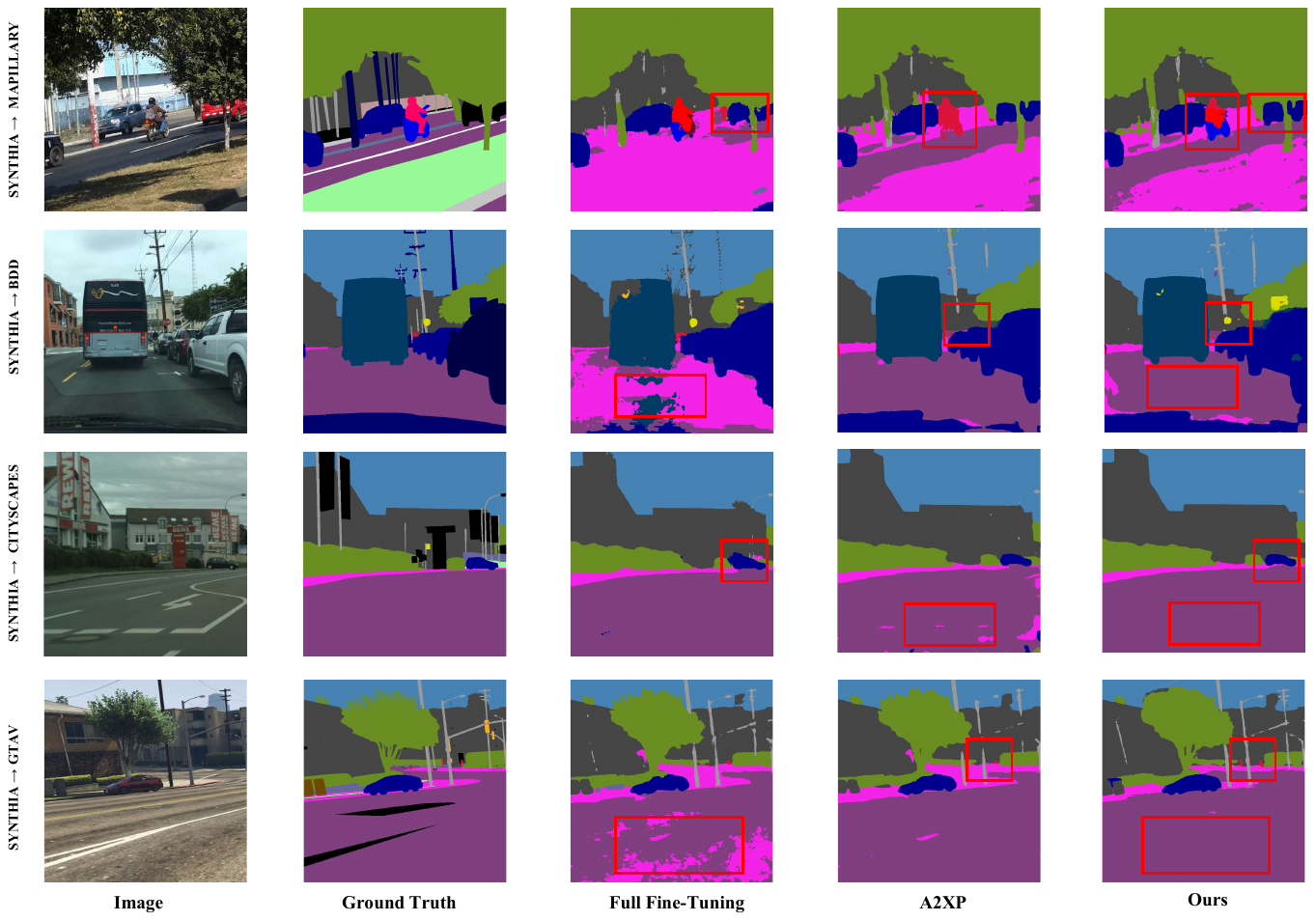}
  \caption{More visual segmentation results under S $\rightarrow$ \{C, B, M, G\} setting. The proposed SAGE is compared with Full Fine-Tuning and A2XP. Compared to alternative methods, SAGE yields less noise and demonstrates a better ability to capture details.}
  \label{fig:more_visual}
\end{figure*}

Table \ref{tab:speed} compares our method with Full Fine-Tuning and A2XP in terms of parameter count, average training time per iteration, and average inference time per image. Our model contains less than 2\% of the parameters of Full Fine-Tuning, demonstrating its lightweight nature and the fact that it requires no extensive computational resources to update the parameters of the pretrained model.
Moreover, our training time is only 35.73 ms longer than that of Full Fine-Tuning, a negligible increase considering the significant reduction in parameters. In addition, our method reduces the training time by 95.53 ms compared to A2XP, another black-box domain generalization method, indicating that the additional steps introduced by our design do not result in a major increase in training time.
However, due to the more complex forward pass, our inference time is slightly longer than that of Full Fine-Tuning and A2XP. This is expected and remains within an acceptable range.
These results demonstrate that our approach achieves substantial parameter savings while maintaining competitive training and inference efficiency.

\subsection{Details of ablation study}

\begin{table}[ht]
\centering
\huge
\caption{Comparison of different prompt generator on mIoU performance.}
\resizebox{1.0\linewidth}{!}{
\begin{tabular}{l|ccccc|ccccc}
\toprule
& \multicolumn{5}{c|}{\textbf{Trained on G}} & \multicolumn{5}{c}{\textbf{Trained on S}} \\ \cline{2-11}
& $\rightarrow$C & $\rightarrow$B & $\rightarrow$M & $\rightarrow$S & Avg. & $\rightarrow$C & $\rightarrow$B & $\rightarrow$M & $\rightarrow$G & Avg. \\
\midrule
Border      & 47.59 & 38.34 & 45.98 & 31.09 & 40.75 & 39.96 & 33.90 & 35.50 & 38.49 & 36.84 \\
A-Border   & 51.38 & 39.35 & 44.12 & 33.50 & \textbf{42.09} & 41.66 & 33.91 & 37.38 & 38.57 & \textbf{37.88} \\
Full    & 46.60 & 40.10 & 43.46 & 32.89 & 40.76 & 42.36 & 33.52 & 34.73 & 36.55 & 36.79 \\
A-Full & 47.26 & 39.14 & 43.02 & 33.77 & 40.80 & 40.87 & 35.91 & 36.29 & 37.26 & 37.58 \\
\bottomrule
\end{tabular}}
\label{tab:SPG_type}
\end{table}

Table \ref{tab:SPG_type} presents the segmentation performance of different generator types under two transfer settings: G $\rightarrow$ \{C, B, M, S\} and S $\rightarrow$ \{C, B, M, G\}. The style-prompt generator used in Section \ref{method} is denoted as A-Border. A-Full is largely similar to A-Border, except that its prompt template shares the same spatial dimensions as the input image. Its modulation network consists of three cascaded ModulatorBlocks followed by a convolutional layer that projects the output to the same number of channels as the input image. This is then upsampled to match the template size, enabling pixel-level reweighting.
However, while A-Full introduces additional information, it may also lead to a greater loss of original image features. As a result, it achieves a lower mIoU, which is 1.29\% lower than A-Border when trained on domain G. Moreover, for both Border and Full variants, their adaptive versions consistently achieve higher mIoU scores. This validates the effectiveness of our modulation network, which adaptively adjusts different prompt regions according to the input image.

\begin{table}[t]
\centering
\caption{Comparison of different initialization strategies on mIoU performance.}
\huge
\resizebox{1.0\linewidth}{!}{
\begin{tabular}{l|ccccc|ccccc}
\toprule
& \multicolumn{5}{c|}{\textbf{Trained on G}} & \multicolumn{5}{c}{\textbf{Trained on S}} \\ \cline{2-11}
& $\rightarrow$C & $\rightarrow$B & $\rightarrow$M & $\rightarrow$S & Avg. & $\rightarrow$C & $\rightarrow$B & $\rightarrow$M & $\rightarrow$G & Avg. \\
\midrule
Zero      & 46.60 & 40.10 & 43.46 & 32.89 & 40.76 & 39.95 & 36.55 & 35.22 & 38.26 & 37.50 \\
Uniform   & 47.49 & 39.66 & 41.59 & 32.79 & 40.38 & 39.95 & 35.59 & 35.39 & 36.61 & 36.89 \\
Normal    & 46.77 & 41.04 & 46.07 & 32.65 & 41.63 & 41.80 & 32.86 & 36.37 & 38.32 & 37.34\\
Meta & 51.38 & 39.35 & 44.12 & 33.50 & 42.09 & 41.66 & 33.91 & 37.38 & 38.57 & 37.88 \\
\bottomrule
\end{tabular}}
\label{tab:SPG_init}
\end{table}
Table \ref{tab:SPG_init} shows how different initialization strategies for the prompt template affect segmentation performance. The Meta initialization involves pretraining on subsets of the source domain for each style using a small number of iterations (200 in our setup) to acquire basic prompting capability, followed by style-specific fine-tuning. As shown, Meta initialization consistently achieves the best mIoU across both transfer scenarios.
In contrast, the Uniform initialization performs the worst in both settings. This is likely because sampling from a uniform distribution over [0, 1] results in a lack of structured patterns in the initial template, making it more prone to converge in suboptimal directions compared to zero or Gaussian initialization.

\begin{table*}[!ht]
\centering
\caption{The sensitivity analysis on style reference. Style abbreviations: L=lakeside, P=pier, V=valley, Vo=volcano, A=ambulance, S=shark, B=barometer, C=can\_opener, Sn=snorkel, T=tennis\_ball.}
\footnotesize
\resizebox{1.0\linewidth}{!}{
\begin{tabular}{@{}l|c|cccc|c|l|c|cccc|c@{}}
\toprule
\multirow{2}{*}{\textbf{Styles}} & \multirow{2}{*}{\textbf{$n$}}
& \multicolumn{5}{c|}{\textbf{Trained on C}} &\multirow{2}{*}{\textbf{Styles}} & \multirow{2}{*}{\textbf{$n$}}
& \multicolumn{5}{|c}{\textbf{Trained on C}}
 \\ \cline{3-7} \cline{10-14}
 &  & $\rightarrow$B & $\rightarrow$M & $\rightarrow$G & $\rightarrow$S & Avg. &  &  & $\rightarrow$B & $\rightarrow$M & $\rightarrow$G & $\rightarrow$S & Avg. \\
\midrule
L,P,V & 3 & 41.46 & 39.39 & 40.86 & 31.34 & 38.26 & B,P,Sn,T,Vo & 5 & 40.04 & 41.04 & 39.78 & 31.25 & 38.03 \\
\textbf{L,P,V,Vo} & 4 & \textbf{45.85} & \textbf{50.27} & \textbf{47.07} & \textbf{32.40} & \textbf{43.90} & B,C,S,Sn,T & 5 & 40.77 & 42.85 & 40.67 & 31.33 & 38.91 \\
A,L,S,V & 4 & 36.98 & 39.60 & 36.46 & 29.36 & 35.60 & L,P,Sn,V,S & 5 & 40.87 & 43.21 & 39.95 & 31.42 & 38.86 \\
P,T,V,Vo & 4 & 41.43 & 40.63 & 39.61 & 31.45 & 38.28 & S,P,B,C,Vo & 5 & 40.65 & 45.28 & 40.26 & 30.61 & 39.20 \\
A,B,C,Vo & 4 & 41.90 & 43.90 & 39.27 & 31.77 & 39.21 & A,C,L,P,V,Vo & 6 & 43.41 & 42.07 & 39.55 & 31.28 & 39.08 \\
A,Sn,T,V & 4 & 39.59 & 43.04 & 38.80 & 30.94 & 38.09 & B,C,L,Sn,T,Vo & 6 & 40.83 & 40.90 & 40.59 & 31.47 & 38.45 \\
\bottomrule
\end{tabular}
}
\label{tab:stylechoice}
\end{table*}

We evaluate the impact of different ImageNet style references and the number of styles $n$ on generalization performance. As shown in Table \ref{tab:stylechoice}, SAGE is robust to various style combinations, with $n=4$ (L, P, V, Vo) achieving the optimal average mIoU of 43.90\%. Notably, increasing $n$ beyond 4 does not yield further significant gains, while inference latency remains independent of $n$ due to our parallelized prompt generation and fusion mechanism.

\subsection{Visual segmentation results}
\label{more_visual}

Figure \ref{fig:more_visual} presents additional segmentation results under the S $\rightarrow$ \{C, B, M, G\} transfer setting, comparing Full Fine-Tuning, A2XP, and our method.
It is evident that Full Fine-Tuning produces more noisy predictions, leading to issues such as overlapping boundaries and incomplete segmentation. For instance, in domains M and C, the car boundaries are poorly defined, and in domains B and G, the road areas contain significant noise.
Although A2XP shows cleaner segmentation with more accurate boundaries, it tends to overlook small objects in the scene. For example, it fails to distinguish between the person and the motorcycle in M, misses the traffic sign in B, and completely ignores a pedestrian in G.
In contrast, our method consistently maintains low-noise segmentation while effectively capturing fine details across all four domains, demonstrating its superior ability to balance boundary clarity with object completeness.

\section{Limitation and future work}
\label{limitation}
Despite the effectiveness of SAGE in generalizing to target models without accessing the internal parameters, architecture, or large-scale training datasets of pre-trained models, several limitations remain. For instance, although our method retains the black-box setting by relying on the backbone’s strong feature extraction capabilities, it still requires access to the final classifier of the target model, with its number of classes adjusted to match the actual dataset. Moreover, the black-box model is required to return not only segmentation outputs but also gradients, which are essential for updating the external style-prompt generator.

To address these limitations, our future work will explore adaptation under fully black-box settings, along with gradient-free optimization strategies for prompt generation. In addition, we plan to extend our method to support domain generalization with multi-modal data and enhance its robustness to out-of-distribution (OOD) samples.

\end{document}